\documentclass[letterpaper]{article} 
\usepackage{aaai2027} 
\nocopyright
\usepackage[hyphens]{url}  
\usepackage{graphicx} 
\def\UrlFont{\rm}  
\usepackage{natbib}  
\usepackage{caption} 
\usepackage{amsmath}
\usepackage{amssymb}
\usepackage{algorithm}
\usepackage{algorithmic}
\usepackage{booktabs}
\usepackage{xr}
\usepackage{xcolor}

\usepackage{newfloat}
\usepackage{listings}
\DeclareCaptionStyle{ruled}{labelfont=normalfont,labelsep=colon,strut=off} 
\floatstyle{ruled}
\newfloat{listing}{tb}{lst}{}
\floatname{listing}{Listing}

\usepackage{booktabs}

\title{Now You See the Hate: Adaptive View Retrieval for Hidden Hateful Illusions}
\author{
    Qianpu Chen\textsuperscript{\rm 1},
    Derya Soydaner\textsuperscript{\rm 1}
}
\affiliations{
    \textsuperscript{\rm 1}LIACS (Leiden Institute of Advanced Computer Science), Leiden University, Leiden, The Netherlands
}

\begin{document}

\maketitle

\begin{abstract}
Hateful optical illusions expose a serious gap in current multimodal safety systems. On original-view hateful illusions, previous work shows that six moderation classifiers achieve at most 20.9--24.5\% accuracy and nine state-of-the-art VLMs remain at or below 10.2\% with illusion-aware prompting, leaving most hidden hate undetected. We formulate 
hidden hateful illusion detection as a perceptual retrieval problem and propose Adaptive View Retrieval. This retrieve-and-calibrate framework assembles a complementary view bank for 
the image and hidden-message templates, adaptively selects which views to trust, retrieves 
hidden-message identities, and calibrates whether the recovered evidence is 
harmful. On HatefulIllusion with a frozen CLIP encoder, Adaptive View Retrieval reaches 93.2\% balanced accuracy on the held-out test split. It substantially outperforms original-view baselines and fixed single-transform filters across hate slangs, hate symbols, and visibility levels. The same design also surpasses official fine-tuned CLIP baselines, matches or exceeds human performance on IllusionMNIST, IllusionFashionMNIST, and IllusionAnimals, and outperforms zoom-out preprocessing on HC-Bench under the SemVink protocol. Together, these results show that robust multimodal moderation requires recovering hidden meaning before deciding whether it is harmful. 
\textcolor{red}{\textbf{Content warning:} This paper contains hateful content and offensive language used solely for scientific research.}

\end{abstract}


\section{Introduction}

\begin{figure}[t]
\centering
\includegraphics[width=0.86\columnwidth]{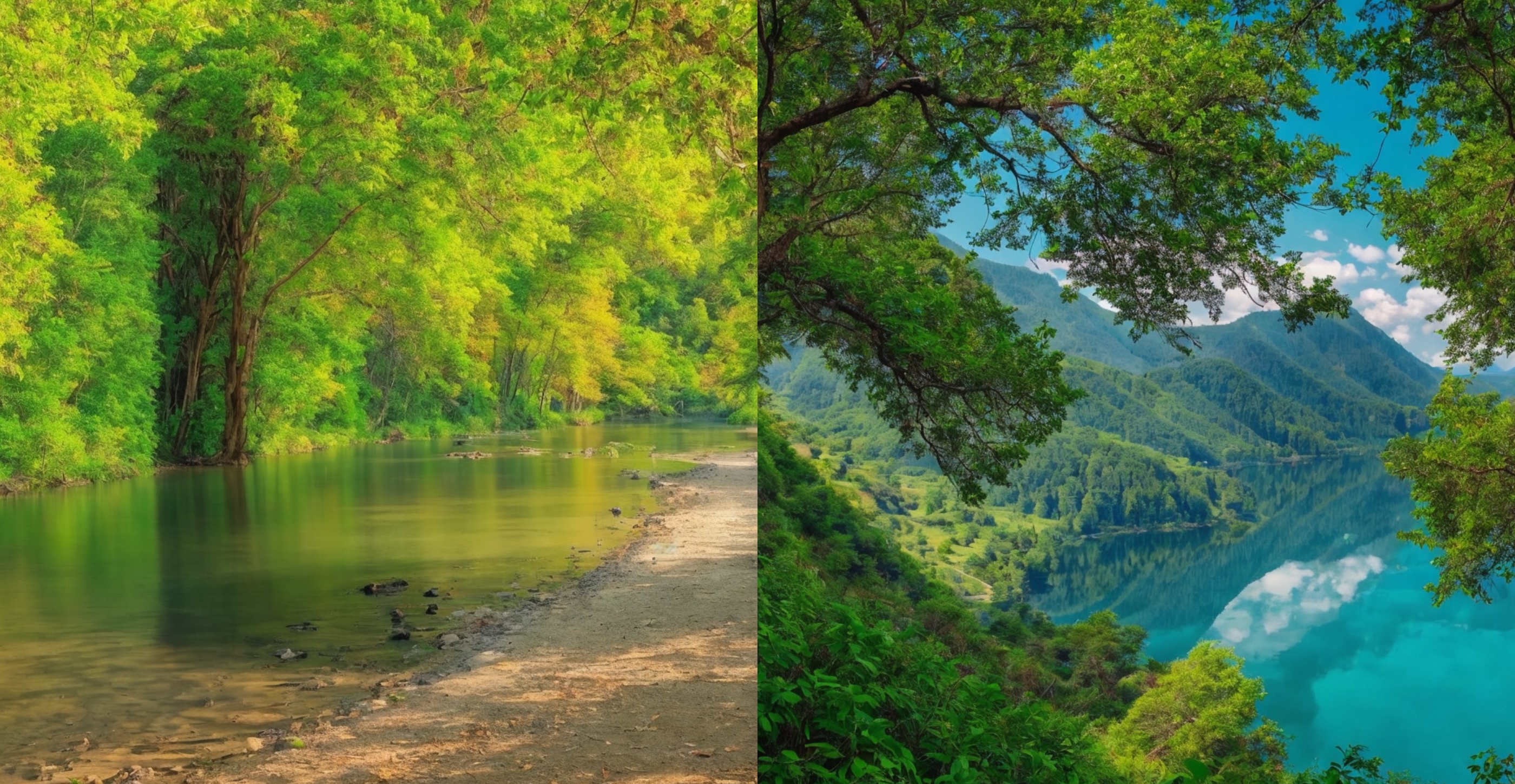}
\caption{The same hidden-shape mechanism can support benign artistic expression or harmful content. \textbf{Left:} a scene that embeds the letter ``L'' as visual art. \textbf{Right:} a scene that uses the same illusion principle to conceal an extremist symbol. Reliable moderation must recover the hidden identity before deciding whether the embedded content is harmful.\protect\footnotemark}
\label{fig:art-vs-hate-illusion}
\end{figure}

Optical illusions are often celebrated as visual art, inviting viewers to discover hidden scenes within ordinary images. Yet the same perceptual trick can also conceal hateful slurs and extremist symbols. Figure~\ref{fig:art-vs-hate-illusion} illustrates this contrast:
the left image embeds the letter ``L'' as an innocuous artistic pattern, while the right image uses the same compositional mechanism to hide a harmful symbol.
For humans, recovery of the hidden scene involves perceptual organization as described by Gestalt principles~\cite{wertheimer1938laws, palmer2002perceptual, wagemans2012century}. For example, closure groups fragmented contours into coherent shapes, and figure-ground organization separates a foreground object from its surrounding scene, revealing the hidden message.

\footnotetext{Both images were generated using the HatefulIllusion ControlNet + Stable Diffusion pipeline~\cite{qu2025hate} with the same scene prompt and seed but different hidden-message control images.}

Generated images are widely shared, yet hateful optical illusions break standard moderation. A benign-looking scene can hide a slang term or extremist symbol in its visual structure, while safety models usually judge only the original view and therefore lack access to the hidden structure.
This \textbf{perceptual access} failure is already severe at scale~\cite{qu2025hate}. The hidden message can remain invisible to models even when humans can read it. On original-view hateful illusions, six moderation classifiers achieve at most 20.9--24.5\% accuracy, and nine state-of-the-art VLMs, including GPT-4o and Gemini-2, remain at or below 10.2\% even with illusion-aware chain-of-thought prompting. With existing moderation pipelines largely failing on this setting, stronger defenses must change \emph{how} the image is observed, not only how the surface scene is classified. Simple transforms can reveal hidden structure, but no single filter is best for every illusion. Slangs, symbols, and visibility levels call for different views.

We argue that recovering hidden hate is fundamentally a \textbf{perceptual retrieval} problem rather than an image classification problem. Our answer is Adaptive View Retrieval. It builds a complementary view bank from 
perceptual-organization operations such as contrast change, contour closure, figure-ground separation, and coarse smoothing. For each image, the model learns which views to trust, applies the same view bank to images and hidden-message templates, and retrieves the best-matching template 
rather than fusing views into a single opaque representation. 

On HatefulIllusion~\cite{qu2025hate}, which contains AI-generated illusions hiding hate slangs and hate symbols, Adaptive View Retrieval reaches 93.2\% balanced accuracy on the held-out message-level test split, well above original-view baselines and fixed filter rules. The same adaptive retrieval design surpasses official fine-tuned CLIP baselines and human 
performance on IllusionMNIST, IllusionFashionMNIST, and IllusionAnimals~\cite{rostamkhani2025illusory}.  
The complementary view bank also substantially improves hidden-content QA on HC-Bench~\cite{li2025semvinkadvancingvlmssemantic}, outperforming both the original image and SemVink-style zoom-out.  
Our contributions are:
\begin{itemize}
    \item We propose Adaptive View Retrieval, a retrieve-and-calibrate architecture that separates hidden-message recovery from moderation calibration through adaptive multi-view template retrieval.
    \item We show that adaptive view selection, not any single classical transform, drives reliable hidden-hate recovery across targets and visibility levels.
    \item We demonstrate generalization beyond hateful-content moderation by surpassing IllusoryVQA fine-tuned CLIP baselines and human performance 
    on IllusionMNIST, IllusionFashionMNIST, and IllusionAnimals. 
    \item Under the official SemVink protocol on HC-Bench, our complementary adaptive multi-view bank substantially improves VLM hidden-content recognition over zoom-out preprocessing on the same backbone.  
\end{itemize}

\section{Related Work}
\textbf{Multimodal Safety and Image Moderation.}
Safety alignment has focused mainly on text-only LMs, but MLLMs face additional risks because decisions depend on both language and visual evidence. Recent work improves MLLM safety through explicit safety reasoning, as in SURE~\cite{gou-etal-2025-sure}. A broader literature highlights unfaithful multimodal behavior: models may produce plausible answers that do not reflect the image, which is 
dangerous for safety-critical use~\cite{chen2026surveymultimodalhallucinationevaluation}. For moderation, this yields a basic requirement: even a well-aligned model cannot judge safely if it cannot access the relevant visual evidence.

We study a stark form of this problem, \emph{perceptual access failure}. Illusion-style images can embed a secondary semantic layer that is hard to recover from the original view. 
Moderation classifiers and VLMs largely miss concealed hate~\cite{qu2025hate}. IllusoryVQA studies fixed filtering as mitigation~\cite{rostamkhani2025illusory}.
Building on these benchmarks, we cast hidden harmful illusion detection as perceptual retrieval and propose adaptive multi-view template retrieval to recover 
hidden identities before moderation.

\noindent\textbf{Visual Illusions and Illusory Recognition.}
Visual illusions, long used to study human perception~\cite{todorovic2020visual}, also deceive convolutional neural networks, though not always in ways that align with human perception \cite{gomez2019convolutional, gomez2020color}. This research has expanded to VLMs, covering classical optical illusions \cite{zhang2023grounding} and related perceptual phenomena \cite{shahgir2024illusionvqa, guan2024hallusionbench, panagopoulou2024evaluating, zhang2025illusionbench, yang2025illusions, rosario2026vision}. VLMs often misclassify illusion-like images and may rely on memorization rather than perceptual reasoning \cite{ullman2024illusion, shinozaki2025large, sun2026vlms}. Illusions have also been generated \cite{burgert2024diffusion, geng2024visual, gomez2025art, hu2026framework} and used to evaluate AI security \cite{ding2025illusioncaptcha}.

More recently, research has extended beyond classical optical illusions to hidden semantic content embedded in images. VLMs struggle to recognize abstract shapes represented as complex arrangements of visual scene elements, despite their ease for human observers \cite{hemmat2024hidden}. IllusoryVQA \cite{rostamkhani2025illusory} embeds objects into images as natural parts of the scene, while HC-Bench~\cite{li2025semvinkadvancingvlmssemantic} introduces hidden text and objects in AI-generated illusion images. 

\noindent\textbf{Perceptual Filtering and Adaptive Multi-View Recognition.}
Multi-view inference is a standard strategy for improving robustness when a single view is insufficient. Test-time augmentation (TTA) aggregates predictions across transformed views, but naive averaging can be suboptimal and may even flip correct predictions to incorrect ones; learning-based aggregation can yield more reliable gains across models and augmentations \cite{shanmugam2021betteraggregationtesttimeaugmentation}. Our setting differs in that views are not generic augmentations for invariance: the goal is to expose latent structure that is invisible in the original image, so the “best” view is highly instance-dependent. Relatedly, Vision Transformers exhibit reduced texture bias and stronger shape-based recognition, 
with increased robustness to occlusions and spatial perturbations \cite{naseer2021intriguingpropertiesvisiontransformers}. These findings support our emphasis on recovering global, shape-like evidence. Together, they motivate our adaptive multi-view framework for hidden-content retrieval and moderation. 



\section{Method}

\begin{figure*}[!t]
    \centering
    \includegraphics[width=0.86\textwidth]{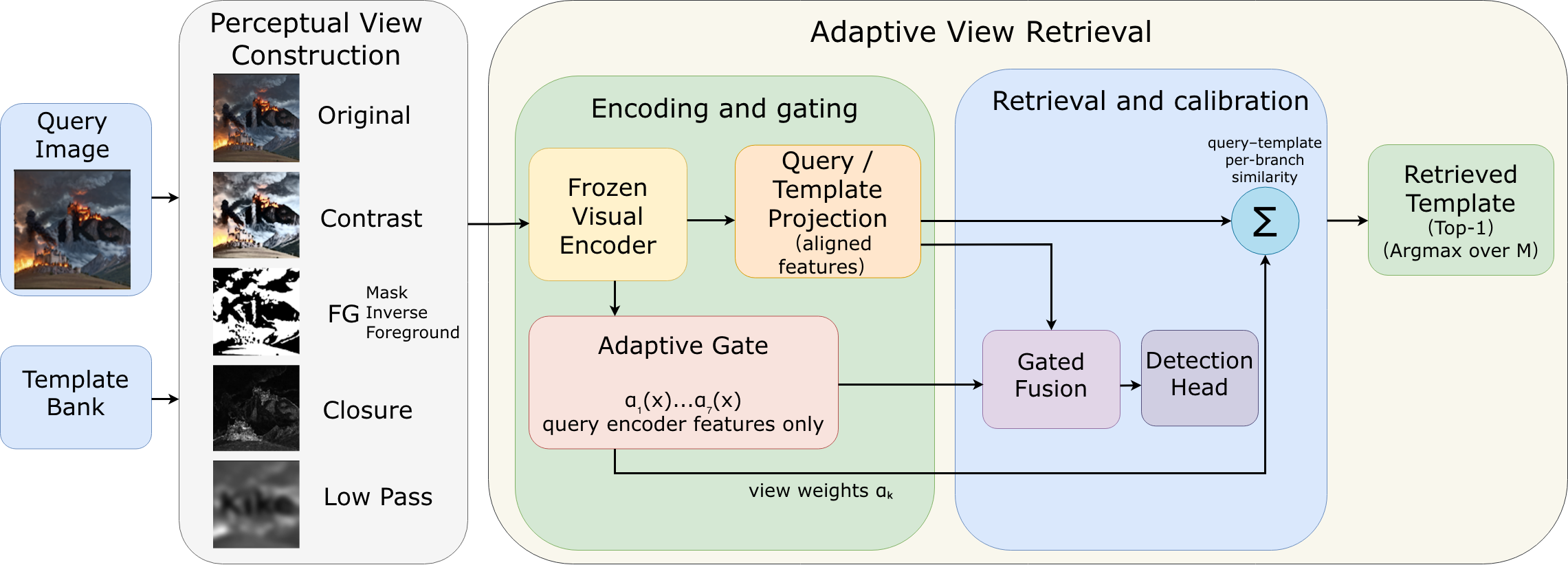}
    \caption{Overview of Adaptive View Retrieval. The template bank stores reference renderings of 
    hidden messages, such as 
    hate slangs and symbols. The same 
    perceptual view bank is applied to the query and every template, so retrieval compares aligned views rather than raw pixels. 
    A frozen CLIP encoder embeds each view, and branch projections align query and template features. 
    An adaptive gate reads the query, assigns weights to each view, combines branch similarities, 
    and feeds a gated representation to a calibration head for the harmful/benign decision. This lets the model trust different views 
    instead of one fixed filter.}
    \label{fig:adaptive-framework}
\end{figure*}

Many images contain hidden content 
that is hard to read from the original view. 
Hidden patterns in illusions often appear only after contrast change, contour emphasis, figure-ground separation, or coarse smoothing, and the most helpful view varies across images. Adaptive View Retrieval addresses this challenge 
with a complementary perceptual view bank: 
deterministic transforms probe different hypotheses about where the hidden signal becomes visible, and the model learns which views 
to trust for each image. The same view bank is applied to the input and to every template, so retrieval compares matched views rather than a single raw image against a single template. Figure~\ref{fig:adaptive-framework} summarizes the full pipeline.

\subsection{Why Retrieval Rather Than Classification?}
\label{sec:why-retrieval}

A fixed $K$-way label head can work on a closed benchmark, but it misstates moderation practice. Operators maintain an expanding \textbf{template library} of known slangs and symbols, and identity should live in that bank rather than in softmax parameters. Retrieval can score newly added templates without retraining the detector head and supplies structured supervision over hidden identities, not only a harmful/benign label. We therefore separate hidden-message recovery from moderation calibration. Retrieval optimizes identity matching, while a calibration head decides whether the recovered evidence counts as harmful. Supplementary Table~1 shows that these roles are complementary: retrieval-only training fails moderation, so reliable moderation requires joint training. 

\subsection{Perceptual View Construction}
\label{sec:view-bank}

Our perceptual view bank combines lightweight, interpretable image operators rather than learned feature extractors. Each operator instantiates a 
transform from classical vision, such as histogram-based contrast enhancement, edge closure, figure-ground segmentation, or coarse low-pass smoothing, but we arrange them as complementary views rather than 
mutually exclusive preprocessing choices. Together they form seven matched views per image: the original view, one contrast-enhanced view, one closure-edge view, three figure-ground views, and one low-pass view. The same deterministic generator is applied to the query image and to every template in the bank, yielding aligned view pairs for retrieval. Because the same view bank is applied to both the input and every template, retrieval compares aligned view pairs in a shared embedding space. 

\textbf{Original View.}
The original view converts the image to RGB and passes it through unchanged. It preserves the surface appearance seen by a standard encoder and serves as the reference against which the derived views are compared.

\textbf{Contrast-Enhanced View.}
It applies a mild Gaussian blur to suppress high-frequency noise, then performs luminance histogram equalization on the Y channel in YCbCr space. The goal is to amplify weak global intensity patterns that may be buried in local texture or uneven lighting.

\textbf{Closure-Edge View.}
The image is converted to grayscale and passed through an edge detector. The edge map is contrast-normalized, closed with a short morphological smoothing step to connect nearby strokes, lightly blurred, and normalized again. The result emphasizes stroke-like boundaries and closed contours that are useful for text- and symbol-like hidden structures.

\textbf{Figure-Ground Views.}
We derive three complementary figure-ground views from the same segmentation procedure. The image is smoothed, contrast-normalized, and thresholded using Otsu's method 
to obtain a 
foreground mask, followed by light morphological cleanup. The \textbf{mask view} preserves the binarized layout.
The \textbf{inverse-mask view} inverts foreground and background. The \textbf{foreground view} composites the original colors onto a neutral background using the mask, isolating the region likely contains the hidden pattern.

\textbf{Low-Pass View.}
This view applies a multi-stage low-pass pipeline 
to suppress local texture while preserving coarse hidden structure. We first apply strong smoothing through Gaussian blur, box blur, and a median filter, then convert the result to grayscale, lightly sharpen 
to recover broad layout, and contrast-normalize the output. The resulting view highlights large-scale patterns 
obscured in the original image.


\subsection{Adaptive View Retrieval}
\label{sec:adaptive-retrieval}

\textbf{Encoding and gating.}
For each branch $k$, a frozen CLIP ViT-B/32 encoder maps query and template views to features $e_x^{(k)}$ and $e_j^{(k)}$, which branch-specific MLPs project to unit-norm embeddings $z_x^{(k)}$ and $z_j^{(k)}$. A gate reads the concatenated pre-projection query features and outputs softmax weights $\alpha_k(x)$ over the bank, assigning mixture coefficients over fixed views rather than computing pairwise interactions between them.\\
\noindent \textbf{Retrieval and calibration.}
Template $m_j$ is scored by a temperature-scaled, gated sum of branch cosine similarities,
\[
S_j(x)=\tau\sum_{k=1}^{K}\alpha_k(x)\,\langle z_x^{(k)}, z_j^{(k)}\rangle,
\qquad
\hat{m}=\arg\max_{m_j\in\mathcal{M}} S_j(x),
\]
where $\mathcal{M}$ is the hidden-message template bank. In parallel, a calibration MLP reads the gated fusion $\bar{z}(x)=\sum_k \alpha_k(x) z_x^{(k)}$ and outputs a harmful/benign logit $d(x)$ without reading the retrieved template directly. Fixed-view and uniform-gate baselines replace $\alpha_k(x)$ with a one-hot rule.

\noindent \textbf{Training and inference.}
\label{sec:joint-training}
We train with a cross-entropy retrieval loss on template scores, a binary cross-entropy calibration loss, and a gate-entropy regularizer to discourage view collapse. On HatefulIllusion, we use loss weights of $1.0$, $0.5$, and $0.01$ respectively, a learning rate of $10^{-3}$, and ten epochs under a message-level 70/15/15 split. Each experiment is repeated with three random seeds (42, 43, 44), and we report mean $\pm$ std over seeds. Only the branch projections, gate, calibration head, and temperature are updated while CLIP remains frozen. The full model reaches 93.2\% $\pm$ 3.1\% balanced accuracy on the held-out test split. We selected the default hyperparameters with a one-at-a-time sweep that varies detection loss ($0.1$, $0.3$, $0.5$, $0.8$), learning rate ($5\times10^{-4}$, $10^{-3}$), gate entropy ($0$, $0.001$, $0.01$), and epoch count ($10$, $15$) while holding the other settings fixed; Supplementary Table~2 shows that balanced accuracy stays near the default. At inference, the model retrieves $\hat{m}$ to recover the hidden message
and outputs $\hat{y}=\mathbb{1}[d(x)\ge 0.5]$ for moderation. On the IllusoryVQA transfer benchmarks, we report top-$k$ retrieval accuracy because the task is hidden-class recognition rather than binary moderation.

\subsection{Datasets}
\label{sec:datasets}

We evaluate Adaptive View Retrieval on three illusion benchmarks. HatefulIllusion is the main moderation benchmark in Experiment~1. IllusoryVQA evaluates generalization to hidden class recognition in Experiment~2. HC-Bench tests multi-view VLM questioning in Experiment~3.

\begin{figure}[h]
\centering
\includegraphics[width=0.8\columnwidth]{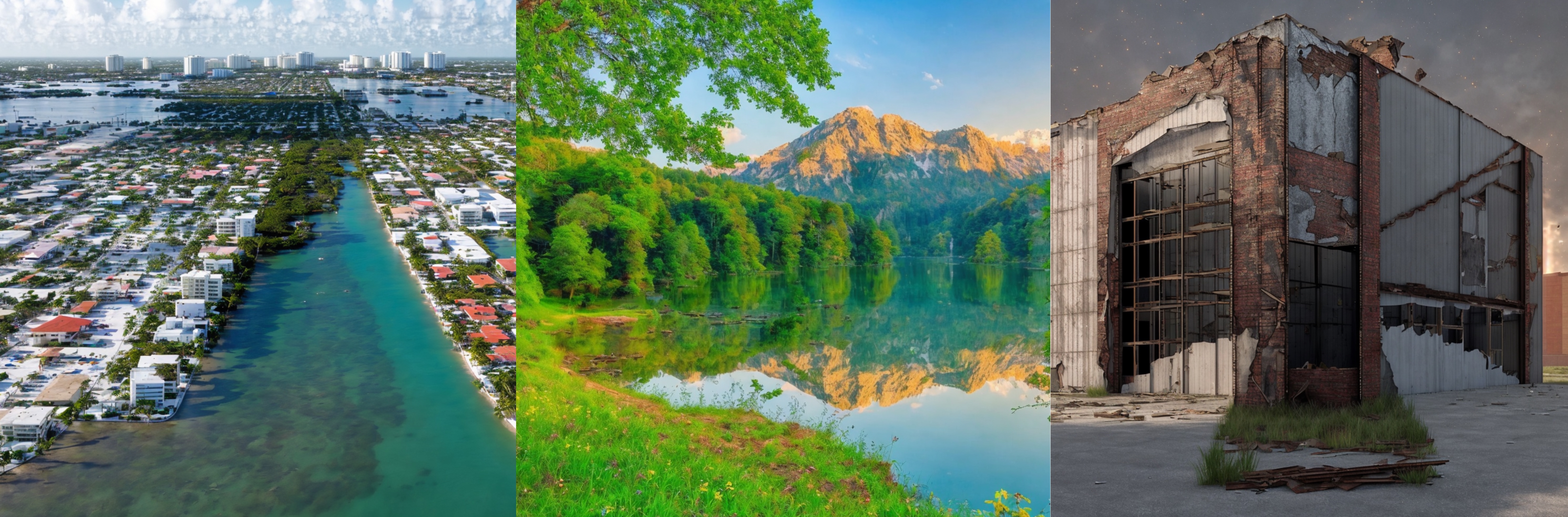}
\caption{Example HatefulIllusion images. From left to right: benign digit, hate slang, and hate symbol illusions. 
}
\label{fig:hatefulillusion-dataset}
\end{figure}

\textbf{HatefulIllusion.}
\label{sec:hatefulillusion-benchmark}
HatefulIllusion~\cite{qu2025hate} contains AI-generated optical illusions that hide hate slang, hate symbols, or benign digit patterns inside natural-looking scenes (Figure~\ref{fig:hatefulillusion-dataset}). Each image has a known hidden-message template and a visibility label that controls how faint the embedded content appears. This benchmark also includes benign digit look-alikes and low-visibility hate examples in which the hidden message is especially faint. Reliable moderation therefore requires both recovering the hidden identity and deciding whether that identity is harmful. Experiment~1 evaluates the full held-out test set and also a hard subset containing all digit look-likes and low-visibility hate examples.

\textbf{IllusoryVQA.}
\label{sec:illusoryvqa-benchmark}
IllusoryVQA~\cite{rostamkhani2025illusory} includes three relevant benchmarks: IllusionMNIST, IllusionFashionMNIST, and IllusionAnimals. Each benchmark embeds a hidden digit, fashion item, or animal into a natural-looking scene generated from an LLM-written prompt. 
The task is to recognize which of ten classes is embedded in the illusion image. Examples are in Supplementary Figure~1. Experiment~2 applies Adaptive View Retrieval with full CLIP fine-tuning and reports Top-1 retrieval accuracy. 

\textbf{HC-Bench.}
\label{sec:hc-bench-benchmark}
HC-Bench~\cite{li2025semvinkadvancingvlmssemantic} contains 112 AI-generated illusion images with hidden content in naturalistic scenes: 56 images with hidden Latin or Chinese text, and 56 with hidden objects (examples in Supplementary Figure~2). Experiments~1 and~2 use CLIP template retrieval, but HC-Bench uses VLM open QA under the official SemVink protocol, including published prompts, zoom-out preprocessing, and scoring rules for text and object cases. Experiment~3 keeps the VLM backbone fixed and tests whether multi-view questioning recovers hidden content better than the original image or SemVink-style zoom-out alone.

\section{Experiments}
We conduct three experiments on complementary illusion benchmarks. Experiment~1 evaluates hidden-hateful-illusion moderation; Experiment~2 tests whether Adaptive View Retrieval 
generalizes beyond hateful content; and Experiment~3 evaluates the complementary view bank on HC-Bench under the official SemVink VLM protocol. Additional Experiment~1 analyses, including component ablations, are in the supplementary material.

\subsection{Experiment 1: Hidden Hateful Illusion Moderation}
\label{sec:benchmark-protocol}

We evaluate hidden-hateful-illusion moderation on HatefulIllusion under the training setup above. We report balanced accuracy on the held-out test set rather than hate-only detection, because reliable moderation must both recover hidden hate and correctly accept benign look-alikes. The supplementary Table~3 compares our method with the moderation classifiers evaluated by Qu et al.~\cite{qu2025hate}.

\noindent \textbf{Do Perceptual Views Help?}
We first ask whether the complementary view bank improves performance and whether any single transform is sufficient.
Figure~\ref{fig:filter-only-adaptive} compares training-free filter baselines, trained original-view retrieval on the original image only, and the full Adaptive View Retrieval model on the mixed hate-and-digit test set. Filter-only baselines remain far below the full model in balanced accuracy, and trained original-view retrieval still trails adaptive multi-view retrieval by a wide margin. This result shows that adaptive multi-view retrieval is necessary for reliable moderation. 

\begin{figure*}[t]
    \centering
    \includegraphics[width=0.81\textwidth]
    {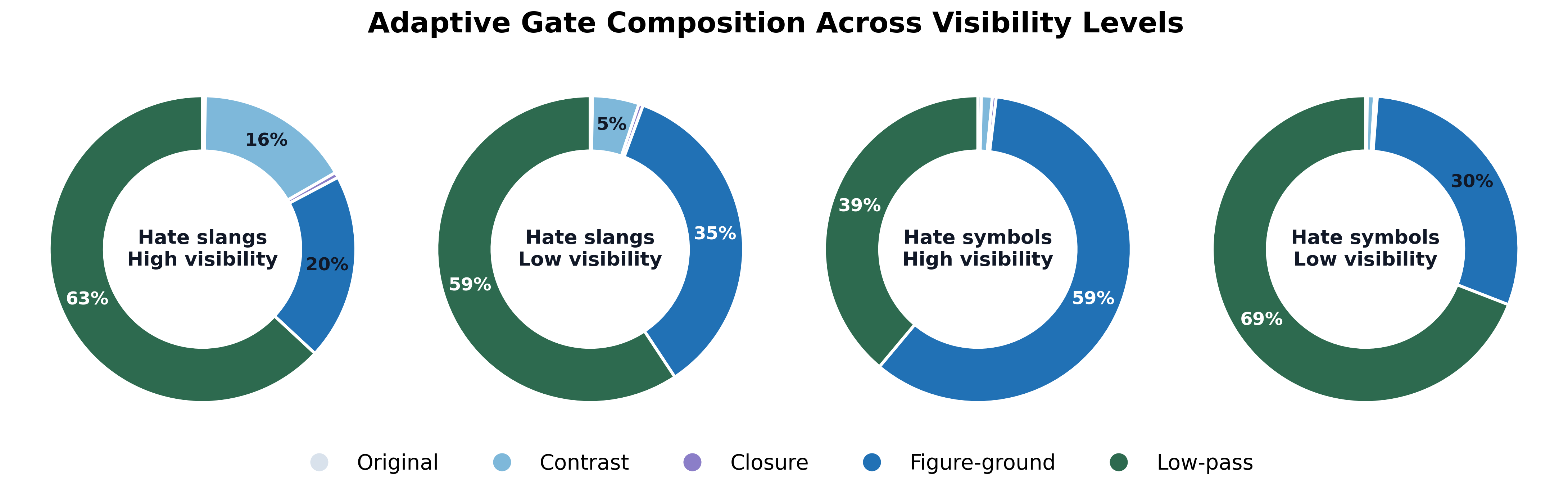}
    \caption{Mean gate composition for harmful targets by target type and visibility.    
    Hate slangs rely predominantly on low-pass evidence. 
    Hate symbols shift toward low-pass filtering as visibility decreases.      
    The gate adapts to both target type and visibility.}   
    \label{fig:gate-composition}
\end{figure*}

\noindent \textbf{Robustness Across Visibility Levels.}
Supplementary Figure~3 summarizes performance across the full test set, the hard subset, and digit specificity.
Filter-only rules reach at most about 60\% balanced accuracy on the full test set and remain weak on the hard subset. Low-pass filtering performs better on digit specificity but remains weak overall, showing that surfacing more hidden signal does not 
make moderation trustworthy.  
Trained original-view retrieval improves to 79.5\%
on the full test set and 78.9\% on the hard subset. Adaptive View Retrieval stays high on all three slices (93.2\%, 92.6\%, and 90.6\% digit specificity). The gap does not shrink on the hard subset, so the full model remains more reliable when the hidden signal is faint and benign digit distractors are included. We next analyze how the gate reallocates weights across visibility levels.

\begin{figure}[h]
    \centering
    \includegraphics[width=1\linewidth]{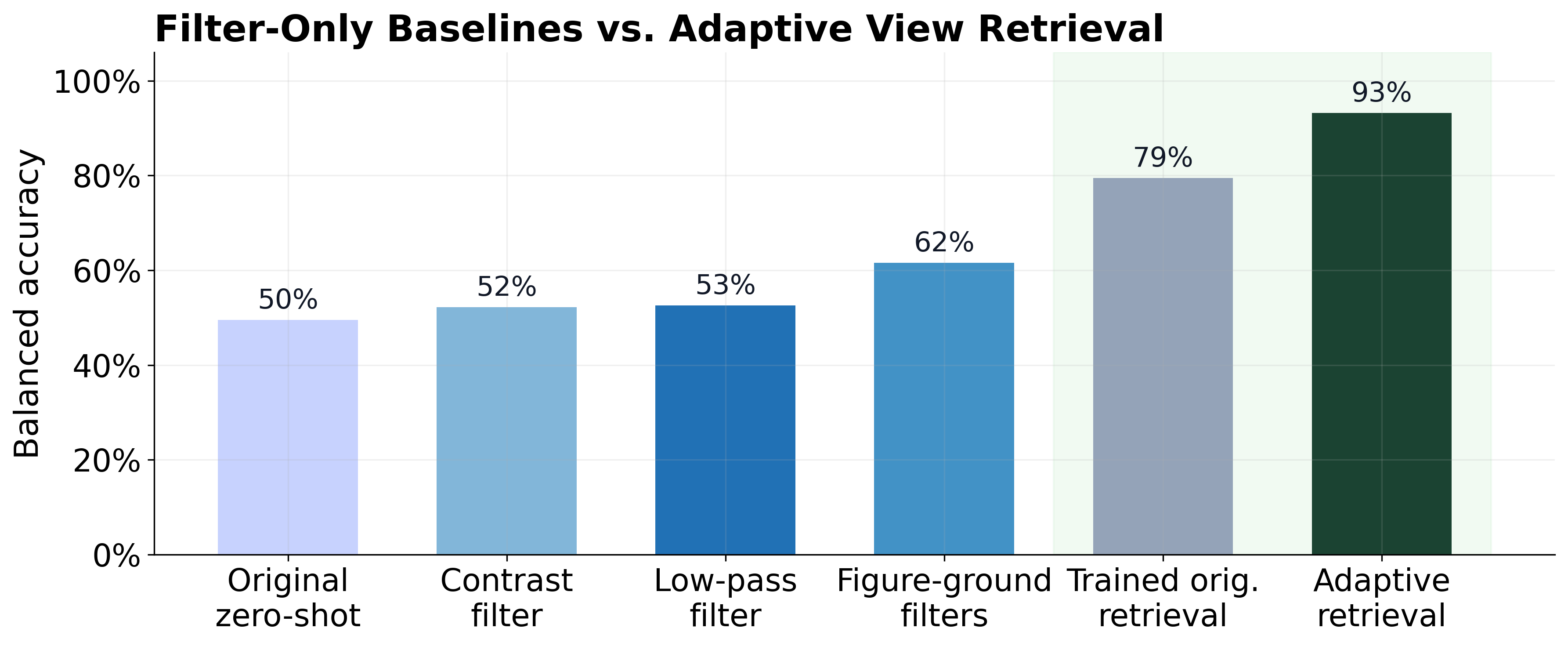}
    \caption{Balanced accuracy on the held-out HatefulIllusion test set. 
    The figure tests whether classic perceptual filters suffice or learned retrieval is required.
    Training-free filters stay near chance, trained original-view retrieval improves to 79.5\%, and Adaptive View Retrieval reaches 93.2\%.}
    \label{fig:filter-only-adaptive}
\end{figure}

\noindent \textbf{What Does Adaptive View Retrieval Learn?}

To understand what the model learns, we analyze both the gate weights over the view bank and the qualitative effect of each perceptual transform. The gate assigns each image a distribution over bank views, so the weights indicate which views the model relies on. We summarize these weights by harmful target and visibility level.

Figure~\ref{fig:gate-composition} shows two complementary patterns. First, the gate learns target-dependent preferences across the view bank. Hate slangs place most weight on low-pass filtering at both visibility levels, whereas hate symbols use a more balanced mixture of low-pass and figure-ground evidence when the hidden signal is clear. Second, visibility modulates these preferences. When the hidden message becomes faint, the gate shifts additional weight toward low-pass filtering for both target types. For hate symbols, figure-ground views receive relatively more weight under high visibility and less under low visibility. This reallocation helps explain why the full model remains more reliable under low visibility (Supplementary Figure~3). Instead of committing to one fixed transform, the model strengthens the bank view best suited to the current visibility. The large low-pass weight does not mean low-pass alone is sufficient. In the filter-only baselines, low-pass remains a weak standalone detector. In the adaptive model, low-pass is one complementary channel inside learned template retrieval, suppressing local texture while figure-ground views contribute shape information.

\begin{figure*}[h]
    \centering
    \includegraphics[width=1\textwidth]
    {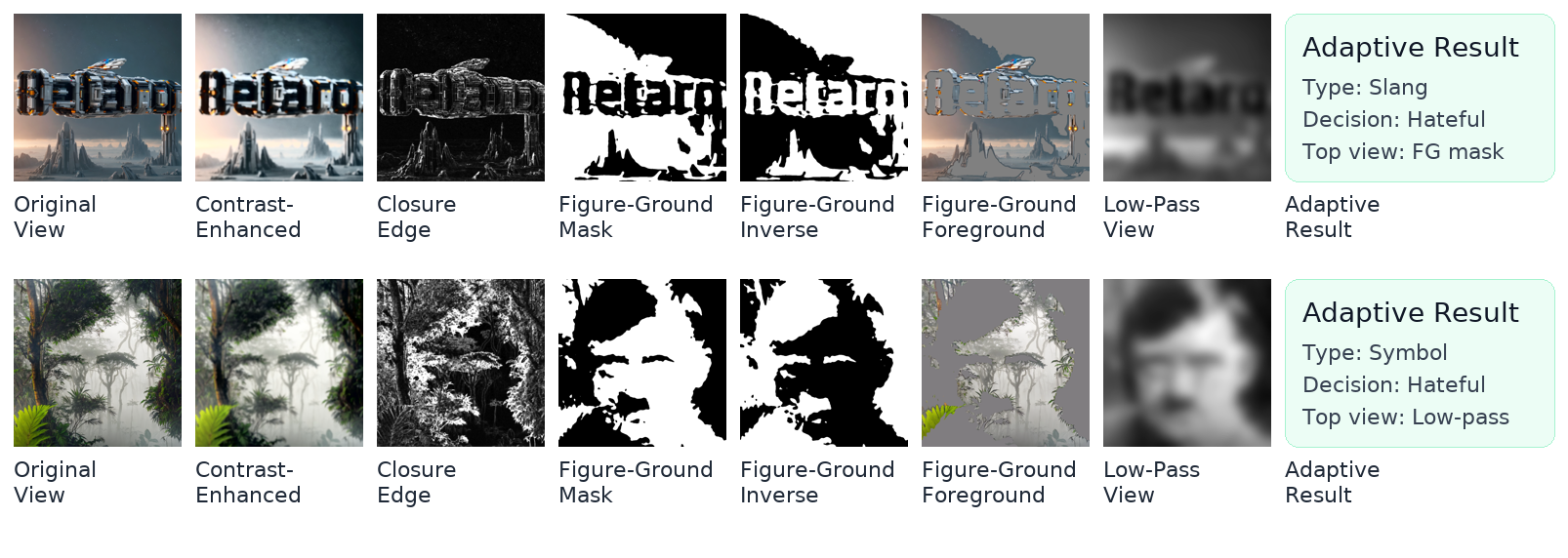}
    \caption{Example hidden-hate images and their complementary view-bank outputs. The dominant view differs across examples,
    illustrating the need for image-dependent view selection. More examples are in Supplementary Figure~4.}
    \label{fig:multiview-gallery}
\end{figure*}

We next examine what each view in the bank contributes in practice. Figure~\ref{fig:multiview-gallery} shows two harmful examples with their transformed views. 
The examples illustrate how each transform exposes different visual evidence and why no single bank view is sufficient.

Contrast enhancement amplifies weak intensity patterns, closure edges emphasize stroke-like structure, figure-ground views separate foreground and background, 
and low-pass filtering reveals broad hidden layouts. Crucially, no single bank view is uniformly best. Some images become clearer under low-pass smoothing, while others are better revealed by figure-ground structure. Because the 
perceptual cue is scene-dependent, fixed single-transform rules behave inconsistently and adaptive weighting over the full bank is necessary.

\noindent \textbf{Failure Analysis.}
\label{sec:failure-analysis}
The model is strong overall, but it still misses harmful content in a small set of test cases. On the held-out test split (1,170 images), Adaptive View Retrieval makes 58 errors: 41 false negatives and 17 false positives on benign digit look-alikes. Most false negatives involve hate symbols, especially when the hidden signal is faint.

\begin{figure*}[t]
\centering
\includegraphics[width=0.9\textwidth]{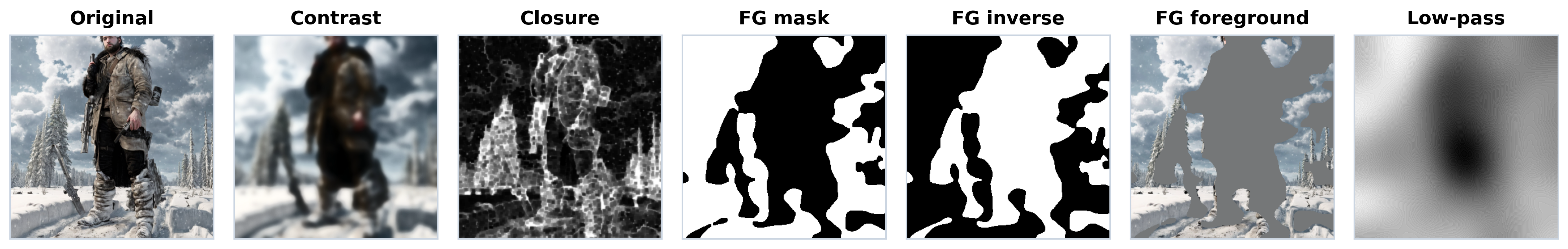}
\caption{A false negative on a low-visibility hate symbol. Contrast and low-pass reveal coarse layout, but figure-ground segmentation fails on this naturalistic carrier, producing a noisy
mask and foreground view. The model over-trusts low-pass evidence, retrieves a digit template instead of the true symbol, and
misses the harmful content.}
\label{fig:failure-case}
\end{figure*}

Figure~\ref{fig:failure-case} shows a 
missed-symbol example. The original scene looks benign, but a hate symbol is embedded at low visibility. The model places the highest gate weight on the low-pass view, yet retrieves a digit candidate and predicts benign with a 
low moderation score. The failure is therefore not an 
absence of hidden-structure evidence, but a breakdown in identity recovery under weak and ambiguous evidence.

Three patterns stand out. First, moderation remains sensitive to template confusion when the hidden signal is faint and benign look-alikes are present in the library. Second, figure-ground views help on many illusions, but deterministic segmentation can fail on complex naturalistic carriers such as textured clothing, snow, and cluttered backgrounds. Third, retrieval and moderation combine weighted evidence from all bank views rather than relying on a single transform. However, the weighted mixture can still fail when the dominant view is ambiguous and complementary views provide noisy or misleading support. These cases motivate human-in-the-loop review and future work on more robust figure-ground extraction or uncertainty-aware gating. 

\subsection{Experiment 2: Generalization on IllusoryVQA}
\label{sec:generalization}

Although our method targets hidden hateful content moderation, the core challenge is perceptual signal recovery. Hidden semantics often appear only under selected transformed views from the bank. We apply the same adaptive retrieval framework to the three IllusoryVQA benchmarks. Table~\ref{tab:illusoryvqa-transfer} reports Top-1 accuracy on illusion images and compares our method with the official IllusoryVQA fine-tuned CLIP baseline, the same model evaluated with the official fixed Gaussian filter, and human annotators.

\begin{table}[h]
\centering
\small
\setlength{\tabcolsep}{5pt}
\begin{tabular}{lccc}
\toprule
\textbf{Method} & \textbf{MNIST} & \textbf{Fashion} & \textbf{Animals} \\
\midrule
Fine-tuned CLIP & 91.80 & 83.90 & 94.36 \\
Fine-tuned CLIP + Fixed Filter & 91.55 & 82.00 & 88.73 \\
Adaptive View Retrieval (Ours) & \textbf{96.0} & \textbf{85.5} & \textbf{97.6} \\
Human & 96.69 & 74.60 & 93.03 \\
\bottomrule
\end{tabular}
\caption{IllusoryVQA transfer. Adaptive View Retrieval outperforms fine-tuned CLIP and CLIP+Fixed Filter on all three splits, matching or exceeding human performance. CLIP baselines and human scores follow Rostamkhani et al.~\cite{rostamkhani2025illusory}, Tables~4 and 2, respectively.}  
\label{tab:illusoryvqa-transfer}
\end{table}
Adaptive View Retrieval surpasses the official fine-tuned CLIP baseline on all three datasets, reaching 96.0\%, 85.5\%, and 97.6\% against 91.80\%, 83.90\%, and 94.36\%. It also outperforms the fixed-filter variant on every split and exceeds human performance 
on IllusionMNIST, IllusionFashionMNIST, and IllusionAnimals. The gain over Fine-tuned CLIP + Fixed Filter is especially clear on IllusionAnimals (97.6\% vs.\ 88.73\%), showing that a single preprocessing rule is not enough even when the backbone is fine-tuned. These results indicate that adaptive view selection generalizes beyond hateful content to broader hidden-class illusion recognition.

\subsection{Experiment 3: Cross-Benchmark on HC-Bench}
\label{sec:hc-bench}

Experiments~1 and~2 study closed-set template retrieval with CLIP. HC-Bench is different. SemVink~\cite{li2025semvinkadvancingvlmssemantic} defines it as VLM open QA over diverse hidden text and objects, with official prompts, zoom-out preprocessing, and answer-matching rules rather than fixed-class retrieval. We therefore follow the SemVink protocol and use Qwen2-VL-2B-Instruct as a fixed backbone, enabling a direct comparison of observation strategies under the same QA setup.  

Within this protocol, we ask whether our complementary perceptual view bank improves hidden-content recognition beyond 
the original image or SemVink-style zoom-out alone. We compare three input strategies: 
the original image, zoom-out to a 64-pixel long edge, and multi-view questioning with the same seven-view bank used in Experiments~1 and~2. The VLM, prompts, and scoring rules stay fixed; only the views change. An example is counted correct if any transformed view yields a valid answer under the official rules.

Table~\ref{tab:hc-bench-vlm} reports the results. Multi-view questioning reaches 98.2\% overall accuracy (100.0\% on objects and 96.4\% on text), compared with 17.0\% for the original image and 2.7\% for zoom-out alone. SemVink reports that zoom-out is highly effective on strong VLMs such as Qwen2-VL-72B-Instruct, often reaching near-perfect accuracy. The same preprocessing fails on our smaller backbone because extreme downsampling leaves too little usable visual signal for the encoder. In contrast, the complementary view bank remains effective under the same official protocol, suggesting that classic perceptual transforms can provide a more robust alternative when resolution downsampling is model-dependent.


\begin{table}[h]
\centering
\small
\setlength{\tabcolsep}{4pt}
\begin{tabular}{lccc}
\toprule
\textbf{Method} & \textbf{Object} & \textbf{Text} & \textbf{Overall} \\
\midrule
Original image & 28.6 & 5.4 & 17.0 \\
Zoom-out (SemVink) & 0.0 & 5.4 & 2.7 \\
Adaptive View Retrieval (Ours) & \textbf{100.0} & \textbf{96.4} & \textbf{98.2} \\
\bottomrule
\end{tabular}
\caption{
HC-Bench hidden-content QA under the official SemVink protocol with Qwen2-VL-2B-Instruct. Adaptive multi-view questioning outperforms both the original image and SemVink-style zoom-out using the same complementary view bank as Experiments~1 and~2. Accuracies (\%) are reported on 56 object and 56 text examples (112 total).} 
\label{tab:hc-bench-vlm}
\end{table}

\section{Discussion}

Our results point to a simple but important distinction for hateful illusion moderation. The main difficulty is 
not whether a model knows what a hateful symbol or slang term means, but whether it can access the hidden message in the first place. Qu et al.~\cite{qu2025hate} show that this access problem is severe for original-view moderation systems and VLMs. Our experiments suggest that the next question is not whether classic perceptual transforms help, but how to organize and adaptively select them across images. 

\textbf{Why Fixed Views Are Not Enough.}
The filter-only baselines make this clear. Several individual transforms from the bank surface part of the hidden signal, and in some cases they improve recall over the original view. Yet their balanced accuracy remains much lower, which means they also introduce false alarms. In other words, a single classic transform can make hidden hate easier to see without making moderation consistently more trustworthy. This matches the broader pattern in our experiments. Different targets respond to different bank views, and a rule that works well on one group can fail on another. Applying one fixed blur, contrast, or figure-ground pipeline therefore helps, but it does not substitute for the full complementary bank with learned selection.

\textbf{What the Gate Learns.}
Adaptive View Retrieval works differently because it does not commit to one bank view in advance. As Figure~\ref{fig:gate-composition} shows, the gate reallocates weight across the complementary view bank as both the harmful target and the visibility level change. When the hidden signal is faint, both hate slangs and hate symbols place more weight on low-pass filtering. When the signal is clearer, especially for symbols, figure-ground views receive 
more weight. This adaptive reallocation helps explain why the full model remains more reliable under low visibility (Supplementary Figure 3) while fixed single-transform rules remain far less trustworthy for moderation (Figure~\ref{fig:filter-only-adaptive}). The 
gallery in Figure~\ref{fig:multiview-gallery} shows the same idea from another angle. Different images become clearest under different views, so image-dependent weighting is not an optional 
feature of our method, but the core requirement of the task.

\textbf{Implications and Limitations.}
The framework is designed for known-message retrieval against a template library, which fits many moderation workflows but not open-vocabulary discovery of unseen hate strings. Performance also drops under stricter target holdout on unseen symbol families, and encoding the full view bank requires more computation than a single encoder pass. 
As Figure~\ref{fig:failure-case} illustrates, figure-ground views can break on complex naturalistic carriers, and weak hidden signals can still lead to template confusion with benign look-alikes. The IllusoryVQA transfer benchmarks and the HC-Bench cross-benchmark provide complementary evidence that the perceptual view bank generalizes beyond hate-specific 
retrieval, although HC-Bench evaluates VLM open QA rather than CLIP retrieval. These limits mark where the approach is most useful. It is best suited to perceptual recovery from a curated candidate set when the most informative observation
changes across images. 

\textbf{Social Impact.}
Hidden hateful illusions expose a real weakness in current moderation systems: harmful content can look benign in the original view while remaining readable to humans. Adaptive View Retrieval addresses this gap by recovering hidden identities through interpretable multi-view evidence rather than a single opaque classifier score. Reviewers can see which view the model trusted and which template it retrieved, making the system easier to audit and better suited to human-in-the-loop review.

The main impact is 
decision support for platforms 
maintaining libraries of known slurs and symbols. New threats can be added as templates without retraining the full stack, and adaptive view selection 
may extend to other hidden-pattern recovery tasks beyond hate moderation. The method should still be used with human oversight, since false positives and adaptive evasion remain possible, but it offers a concrete step toward safer and more explainable hidden-hate detection.

\section{Conclusion}

Hateful optical illusions expose a blind spot in current moderation systems. Harmful content can be present in an image while remaining difficult to recover from the original view alone. We address this problem with Adaptive View Retrieval, which organizes classic perceptual transforms into a complementary view bank for both images and candidate templates, then learns which views to trust for each example. On HatefulIllusion, the 
model reaches 93.2\% balanced accuracy on the held-out test split, well above original-view baselines and fixed single-transform filters across target types and visibility levels. The perceptual view bank also generalizes to IllusoryVQA 
and HC-Bench VLM hidden-content QA under the 
SemVink protocol. Together, these results show that robust multimodal moderation requires recovering hidden meaning before deciding whether it is harmful.
\clearpage
\bibliography{aaai2027}

\clearpage
\appendix

\twocolumn[
\begin{center}
{\LARGE\bfseries Supplementary Material:\\
Now You See the Hate: Adaptive Multi-View Retrieval for Hidden Hateful Illusions\par}
\vspace{1.5em}
{\large
Qianpu Chen\textsuperscript{\rm 1},
Derya Soydaner\textsuperscript{\rm 1}\par}
\vspace{0.5em}
{\normalsize
\textsuperscript{\rm 1}LIACS (Leiden Institute of Advanced Computer Science), Leiden University, Leiden, The Netherlands\par}
\vspace{2em}
\end{center}
]

\section{IllusoryVQA and HC-Bench Examples}

\begin{figure}[h]
\centering
\includegraphics[width=1.0\columnwidth]{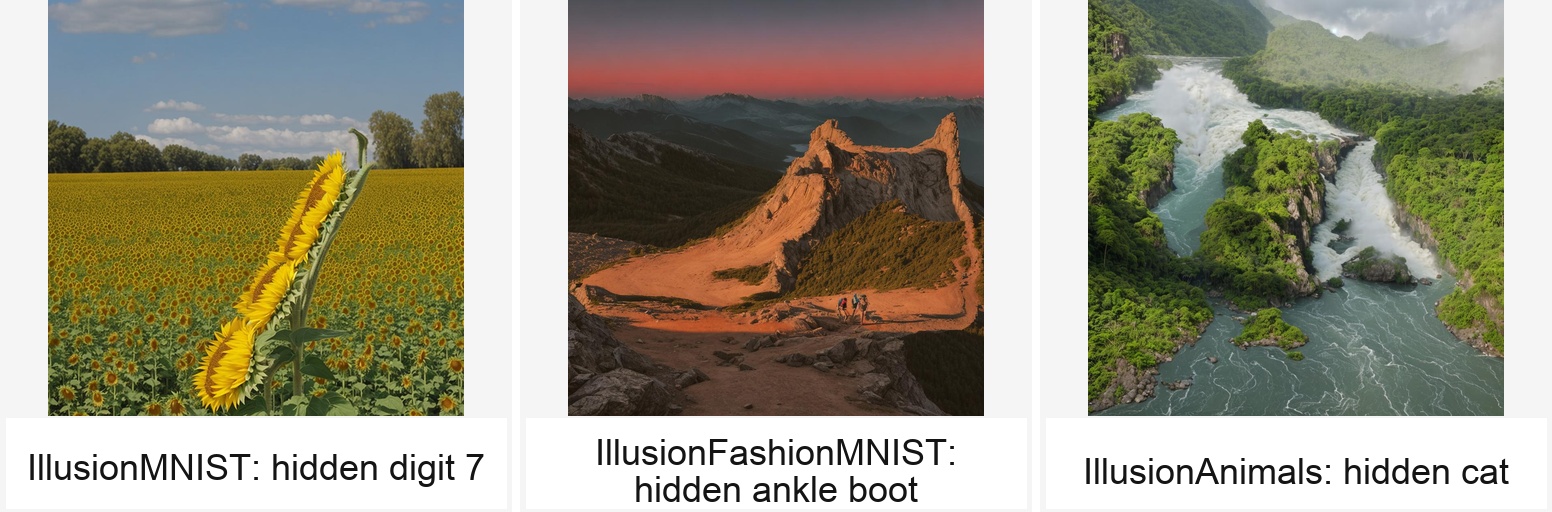}
\caption{Example IllusoryVQA images from IllusionMNIST, IllusionFashionMNIST, and IllusionAnimals. Each scene hides a digit, fashion item, or animal within a natural-looking image.}
\label{fig:illusoryvqa-dataset}
\end{figure}

\begin{figure}[h]
\centering
\includegraphics[width=1.0\columnwidth]{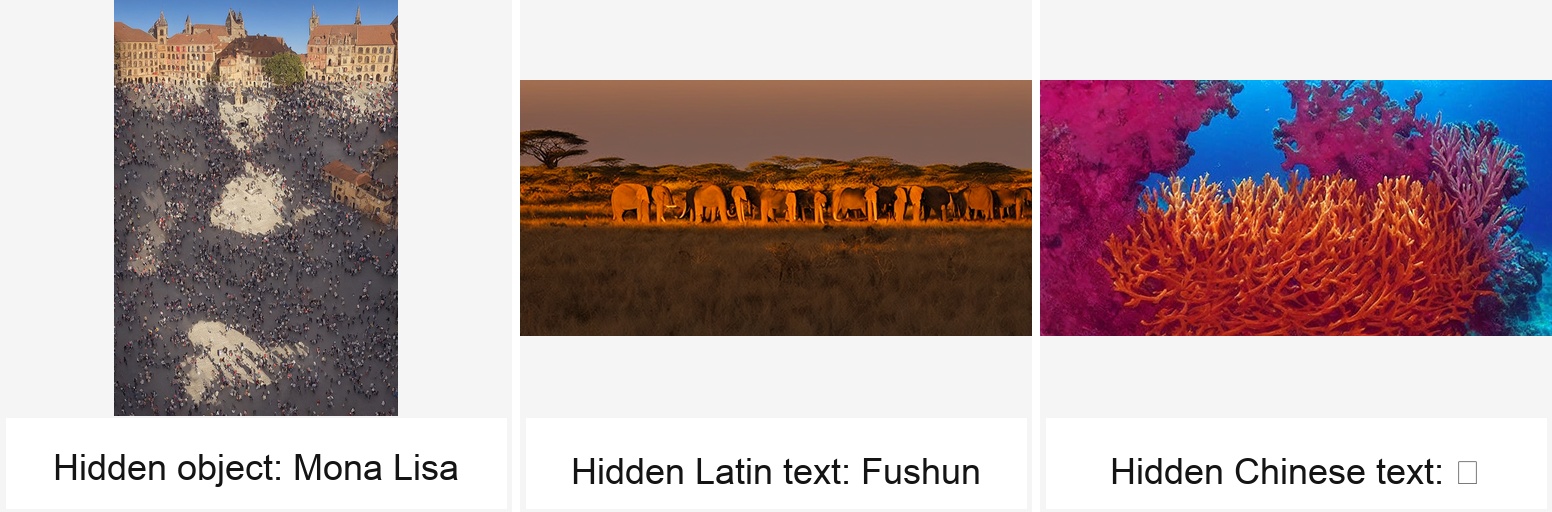}
\caption{Example HC-Bench images with hidden objects, hidden Latin text, and hidden Chinese text embedded in naturalistic scenes.}
\label{fig:hc-bench-dataset}
\end{figure}

The main paper presents HatefulIllusion examples only. Figure~\ref{fig:illusoryvqa-dataset} and Figure~\ref{fig:hc-bench-dataset} introduce the remaining two benchmarks used in our evaluation. IllusoryVQA~\cite{rostamkhani2025illusory} embeds hidden object classes in natural scenes, whereas HC-Bench~\cite{li2025semvinkadvancingvlmssemantic} embeds hidden Latin or Chinese text and hidden objects in AI-generated illusion images. 

\section{Component Ablation}
\label{sec:component-ablation}

We ablate Adaptive View Retrieval under the Experiment~1 protocol from the main paper. Table~\ref{tab:component-ablation} disables one design choice at a time and reports moderation balanced accuracy on the held-out test set.

Three patterns stand out. The complementary view bank matters: using only the original view lowers balanced accuracy from 93.2\% to 79.5\%. Retrieval and calibration play complementary roles: training with retrieval loss only collapses moderation to 47.8\% balanced accuracy because the calibration head receives no harmful/benign supervision. Moderation therefore requires joint training, not retrieval alone. Adaptive gating is important: fixing the gate on the original view drops moderation to 82.6\%, while uniform weighting recovers much of the gain (89.1\%) but still trails the full adaptive model (93.2\%). Together, these results support Adaptive View Retrieval as the shared perceptual mechanism, while retrieval and evidence calibration play complementary roles.

\begin{table}[h]
\centering
\small
\setlength{\tabcolsep}{4pt}
\begin{tabular}{lc}
\toprule
\textbf{Configuration} & \textbf{Moderation (Bal. Acc.)} \\
\midrule
Full model & 93.2 \\
Original view only (no view bank) & 79.5 \\
Retrieval loss only (no calibration) & 47.8 \\
Fixed gate: original view only & 82.6 \\
Fixed gate: uniform weights & 89.1 \\
\bottomrule
\end{tabular}
\caption{Component ablation on HatefulIllusion (test, 3 seeds). Results correspond to Experiment~1 in the main paper. Each row changes one component of the full model while leaving the rest intact. Moderation balanced accuracy measures harmful/benign decisions from the calibration head.}
\label{tab:component-ablation}
\end{table}

\section{Hyperparameter Sensitivity}

We run a one-at-a-time hyperparameter sweep on the main HatefulIllusion model under the Experiment~1 protocol (frozen CLIP ViT-B/32, full view bank, 3 seeds, message-level 70/15/15 split). Table~\ref{tab:hp-sweep} changes one hyperparameter at a time relative to the paper default ($\lambda_{\mathrm{det}}=0.5$, $\lambda_{\mathrm{ent}}=0.01$, learning rate $10^{-3}$, 10 epochs).

\begin{table}[]
\centering
\small
\setlength{\tabcolsep}{4pt}
\begin{tabular}{lcccc}
\toprule
\textbf{Setting} & \textbf{Bal.\ Acc.} & \textbf{Recall} & \textbf{Spec.} & \textbf{Cat.\ Match} \\
\midrule
Paper default & 93.2 $\pm$ 3.1 & 95.9 & 90.6 & 89.3 \\
detection loss $=0.1$ & 92.6 $\pm$ 3.1 & 96.3 & 88.9 & 89.7 \\
detection loss $=0.3$ & 93.0 $\pm$ 2.1 & 96.0 & 90.0 & 89.1 \\
detection loss $=0.8$ & 91.6 $\pm$ 4.2 & 95.4 & 87.8 & 90.9 \\
learning rate $=5\times10^{-4}$ & 93.8 $\pm$ 4.1 & 96.5 & 91.1 & 90.0 \\
gate entropy $=0$ & 91.6 $\pm$ 3.6 & 95.5 & 87.8 & 89.0 \\
gate entropy $=0.001$ & 90.6 $\pm$ 3.6 & 96.7 & 84.4 & 88.4 \\
epochs $=15$ & 92.2 $\pm$ 4.4 & 97.2 & 87.2 & 89.6 \\
\bottomrule
\end{tabular}
\caption{One-at-a-time hyperparameter sweep (mean $\pm$ std over 3 seeds) on the Experiment~1 test split.}
\label{tab:hp-sweep}
\end{table}

\begin{figure*}[t]
\centering
\includegraphics[width=0.91\textwidth]{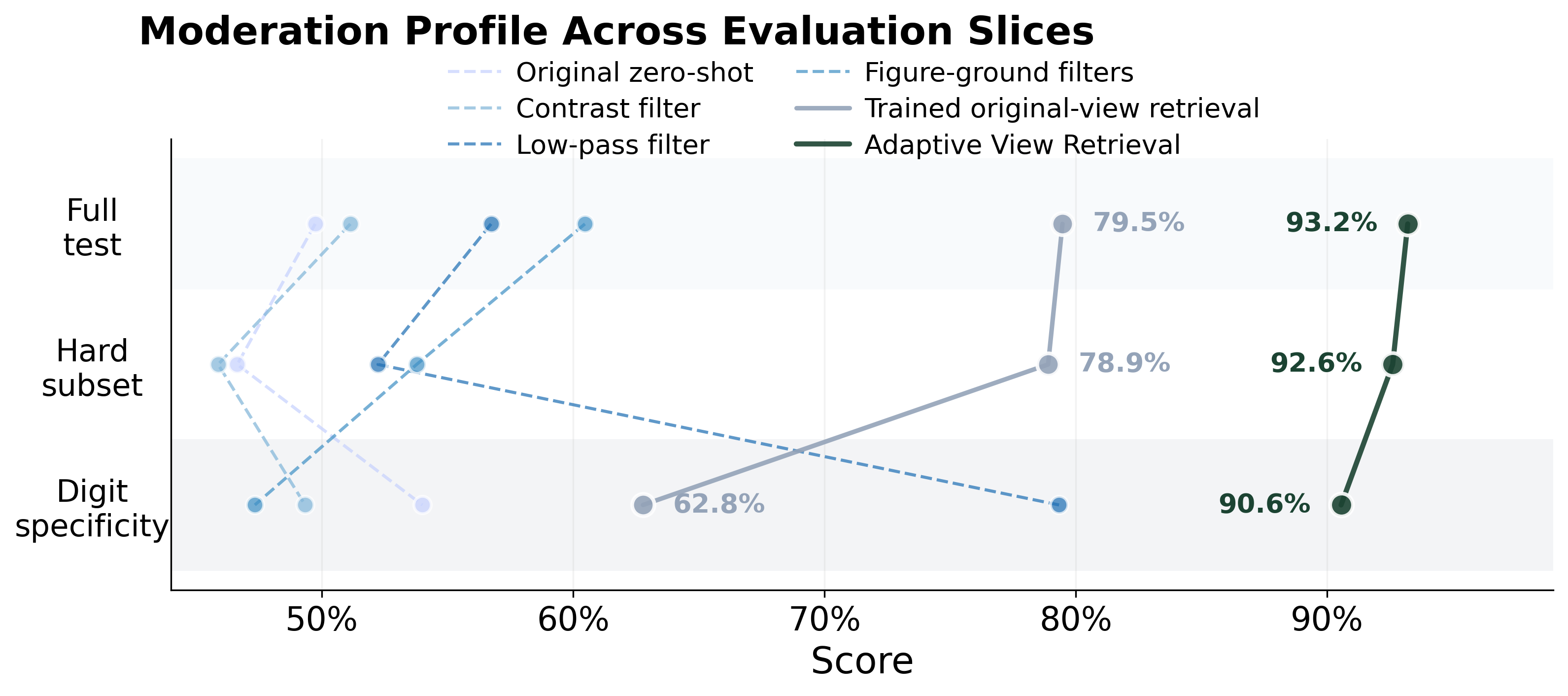}
\caption{Moderation profile on the held-out HatefulIllusion test set. Each line reports one method on the full test set, the hard subset, and benign digits.}
\label{fig:moderation-profile}
\end{figure*}

Balanced accuracy stays near 93.2\% for all variants, and no setting improves on the default by more than 0.6 points. We therefore keep the default hyperparameters used in the main paper.

\section{Comparison with Qu et al.\ Moderation Classifiers}
\label{sec:qu-comparison}

Table~\ref{tab:qu-moderation-comparison} follows the layout of Table 2 from Qu et al. \cite{qu2025hate} and adds Adaptive View Retrieval on the same illusion slices. Qu et al.\ (2025) report zero-shot moderation classifiers on overt messages and hateful illusions at high and low visibility. Our row reports hate detection on held-out hateful-illusion images only, without benign digit distractors, averaged over three seeds. The 93.2\% balanced accuracy reported in the main paper is measured on the mixed test set and is therefore not shown here.   

\begin{table}[]
\centering
\scriptsize
\setlength{\tabcolsep}{3pt}
\begin{tabular}{lccccccccc}
\toprule
& \multicolumn{2}{c}{\textbf{Message}} & \multicolumn{3}{c}{\textbf{Hate Speech Illusions}} & \multicolumn{3}{c}{\textbf{Hate Symbol Illusions}} \\
\cmidrule(lr){2-3}\cmidrule(lr){4-6}\cmidrule(lr){7-9}
\textbf{Method} & Msg & Msg & High & Low & Agg & High & Low & Agg \\
\midrule
\multicolumn{9}{l}{\textit{Off-the-shelf moderation classifiers (Qu et al.)}} \\
Omni & 0.043 & 0.026 & 0.012 & 0.006 & 0.008 & 0.009 & 0.016 & 0.010 \\
SafeSearch & 0.000 & 0.154 & 0.004 & 0.014 & 0.010 & 0.015 & 0.027 & 0.018 \\
Moderation API & 0.000 & 0.128 & 0.000 & 0.000 & 0.000 & 0.011 & 0.000 & 0.009 \\
M-Moderation API & 0.217 & 0.436 & 0.004 & 0.000 & 0.002 & 0.002 & 0.000 & 0.002 \\
Safety Checker & 0.652 & 0.231 & 0.069 & 0.008 & 0.033 & 0.042 & 0.011 & 0.036 \\
Q16 & 0.696 & 0.846 & 0.234 & 0.191 & 0.209 & 0.190 & 0.245 & 0.245 \\
\midrule
\multicolumn{9}{l}{\textit{Ours}} \\
Adaptive View Retrieval & --- & --- & 1.000 & 0.989 & 0.994 & 0.980 & 0.903 & 0.964 \\
\bottomrule
\end{tabular}
\caption{Hate detection on overt messages and hateful illusions, following Qu et al.(2025)\ Table~2. Qu et al.(2025)\ results are from their paper. Our illusion results use held-out hateful-illusion images only. \textit{Agg} denotes the sample-weighted average of High and Low visibility.}
\label{tab:qu-moderation-comparison}
\end{table}

Off-the-shelf classifiers remain near chance on hateful illusions, while Adaptive View Retrieval reaches 0.994 on hate-speech illusions and 0.964 on hate-symbol illusions. This hate-only comparison complements the mixed-test balanced accuracy in the main paper.

\section{Extended Moderation Profile}

Figure~\ref{fig:moderation-profile} shows the full Experiment~1 moderation profile. Each method is scored on the full mixed test set, a hard subset with all digit look-alikes and low-visibility hate, and benign digits alone. Filter-only rules stay weak overall, trained original-view retrieval improves but still lags on the harder slices, and Adaptive View Retrieval remains strongest throughout (93.2\%, 92.6\%, and 90.6\%).

\section{Additional Multi-View Gallery Examples}

\begin{figure*}[t]
\centering
\includegraphics[width=1\textwidth]{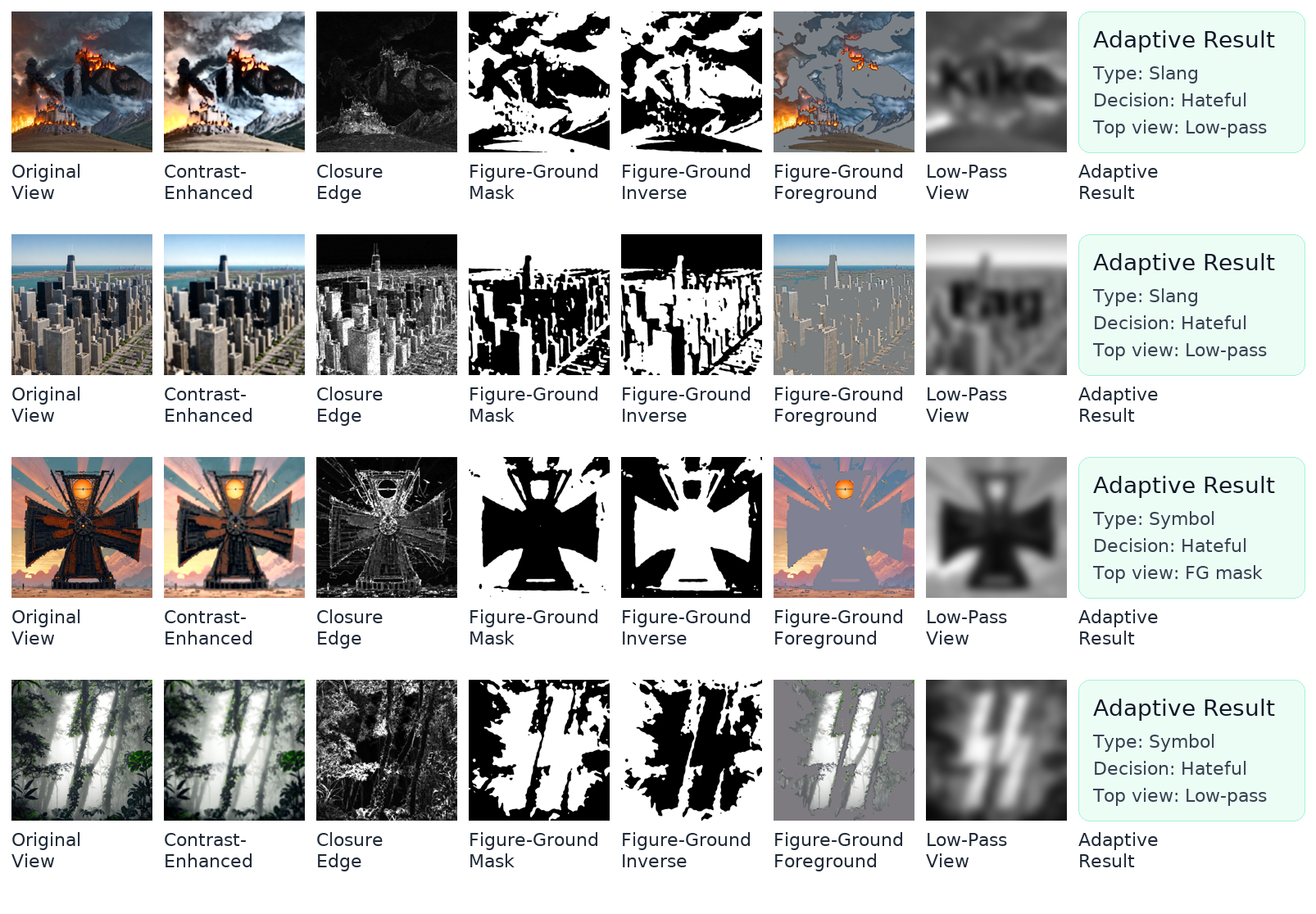}
\caption{Additional multi-view gallery examples on HatefulIllusion. Each row shows one hidden-hate image under the complementary view bank, together with the adaptive result. Contrast enhancement, closure edges, figure-ground masks, foreground extraction, and low-pass filtering expose different aspects of the same image, and the dominant view varies across cases.}
\label{fig:multiview-gallery-supp}
\end{figure*}

The main paper shows two harmful examples and their complementary view-bank outputs. Figure~\ref{fig:multiview-gallery-supp} provides four additional HatefulIllusion cases (two hate slangs and two hate symbols). Each row shows the original image, the six complementary transforms in the view bank, and the adaptive decision. As in the main paper, the clearest view differs across examples: some cases favor low-pass filtering, while others are revealed more clearly under figure-ground separation. These additional examples reinforce the need for image-dependent selection over the full view bank.

\section{Implementation and Reproducibility}

Experiments were run in Python~3.9+ with PyTorch, OpenCLIP ViT-B/32, and Hugging Face \texttt{datasets}/\texttt{transformers}. Training and evaluation used CUDA or Apple MPS when available. HC-Bench VLM experiments used Qwen2-VL-2B-Instruct locally through \texttt{transformers}. All experiment scripts, configuration files, and a module README are included in the public code release under \texttt{experiments/illusion\_perception\_benchmark/}.


\end{document}